\documentclass[10pt,twocolumn,letterpaper]{article}

\usepackage[pagenumbers]{cvpr} 

\usepackage{graphicx}
\usepackage{amsmath}
\usepackage{amssymb}
\usepackage{booktabs}
\usepackage{multirow}

\usepackage[pagebackref,breaklinks,colorlinks]{hyperref}

\usepackage[capitalize]{cleveref}
\crefname{section}{Sec.}{Secs.}
\Crefname{section}{Section}{Sections}
\Crefname{table}{Table}{Tables}
\crefname{table}{Tab.}{Tabs.}

\def\cvprPaperID{*****} 
\def\confName{CVPR}
\def\confYear{2022}

\begin{document}

\title{Scene Clustering Based Pseudo-labeling Strategy for Multi-modal Aerial View Object Classification}

\author{Jun Yu$^1$ , Hao Chang\footnotemark[1] $^1$, Keda Lu$^1$$^2$, Liwen Zhang$^1$, Shenshen Du$^1$, Zhong Zhang$^3$\\
$^1$University of Science and Technology of China,$^2$ Ping An Technology Co., Ltd,\\
$^3$Hefei ZhanDa Intelligence Technology Co., Ltd\\
{\tt\small harryjun@ustc.edu.cn, changhaoustc@mail.ustc.edu.cn, wujiekd666@gmail.com,}\\
{\tt\small zlw1113@mail.ustc.edu.cn, nibility163@163.com, zhangzhong@zalend.com}
}

\maketitle
\renewcommand{\thefootnote}{\fnsymbol{footnote}}
\footnotetext[1]{Corresponding author.}

\begin{abstract}

Multi-modal aerial view object classification (MAVOC) in Automatic target recognition (ATR), although an important and challenging problem, has been under studied. This paper firstly finds that fine-grained data, class imbalance and various shooting conditions preclude the representational ability of general image classification. Moreover, the MAVOC dataset has scene aggregation characteristics. By exploiting these properties, we propose Scene Clustering Based Pseudo-labeling Strategy (SCP-Label), a simple yet effective method to employ in post-processing. The SCP-Label brings greater accuracy by assigning the same label to objects within the same scene while also mitigating bias and confusion with model ensembles. Its performance surpasses the official baseline by a large margin of +20.57\% Accuracy on Track 1 (SAR), and +31.86\% Accuracy on Track 2 (SAR+EO), demonstrating the potential of SCP-Label as post-processing. Finally, we \textbf{win the championship both on Track1 and Track2} in the CVPR 2022 Perception Beyond the Visible Spectrum (PBVS) Workshop MAVOC Challenge~\cite{mavoc}. Our code is available at \url{https://github.com/HowieChangchn/SCP-Label}.

\end{abstract}

\section{Introduction}
\label{sec1}

With the advent of research on computer vision, the performance of ATR has witnessed incredible progress. Synthetic aperture radar (SAR) and electro-optical (EO), two popular modalities for modern remote sensing (RS) systems, have the advantage of capturing visible light features and ignoring climate effects, respectively. Although each individual data modality has advantages and disadvantages, traditional RS systems typically leverage only a single modality. Therefore, ATR algorithms that utilize multiple data modalities may be able to mitigate the disadvantages associated with each sensor type.

\begin{figure}[t]
	\centering
	\includegraphics[width=0.45\textwidth]{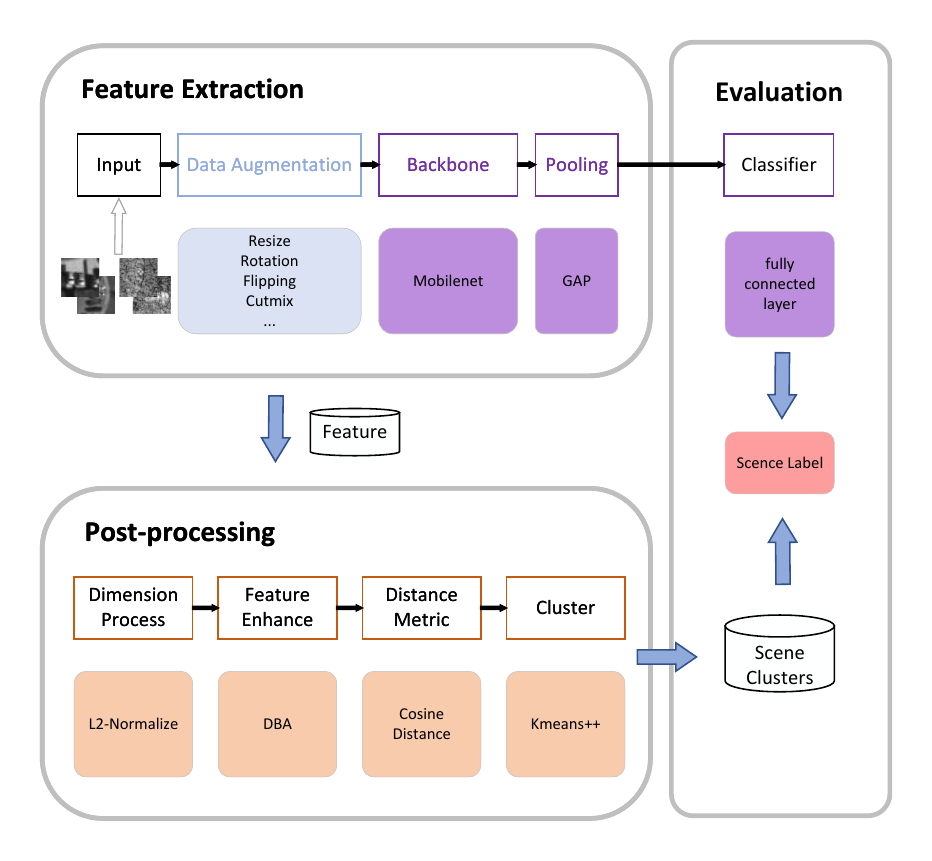} 
	\caption{{\bf SCP-Label for MAVOC.} We utilize a convergent model in the supervised training step for feature extraction, followed by a pseudo-label strategy based on scene clustering for post-processing.}
	\label{fig_intro}
\end{figure}

A lot of research has been done on ATR for both EO and SAR images. In terms of EO-ATR, a large number of related research results have been achieved due to the use of early traditional visual methods~\cite{arguedas2015texture} and current convolutional neural networks (CNNs)~\cite{rice2018convolutional} to extract features. In terms of SAR-ATR, the acquisition of SAR image data is difficult, and related research is lagging behind. Recently, there are also some CNNs structures designed for SAR images, such as fuzzy logic and dynamic neural networks~\cite{tzeng1998fuzzy}, and end-to-end CNNs~\cite{chen2014sar} for SAR images. But most research directions focus on their respective fields, i.e. ATR development for SAR and EO are disjoint.

In contrast, ATR on multiple data modalities has been under studied. In fact, the area for aerial view images poses further challenges in ATR with lower target resolution, different target texture and light reflectance, high inter-class similarity, etc. In the general ATR algorithm, the image is extracted by CNNs and classified by a classifier. However, if the identified objects are fine-grained and affected by the factors mentioned above, and what is worse, the dataset has a long-tailed distribution, the classification results will be confused and biased towards the majority class. Both class imbalance and fine-grained multi-modal data has not been thoroughly evaluated in most existing ATR algorithms.

As a preliminary of our work, we investigate the performance of general image classification algorithm in the context of class-imbalanced and fine-grained MAVOC dataset. Specifically, to figure out how algorithms perform, we manually divide the training set and calculate the bias and confusion of the model in each category. By conducting validation experiments, we find that resampling helps balance the classification performance of the model and observes confusing categories in the dataset. More importantly, our method is motivated by the further observation that, despite the aforementioned facts, the MAVOC dataset has scene aggregation characteristics. The model will recognize objects in the same scene as different categories due to the interference of various factors.

Therefore, in this paper, for exhaustively improving the recognition performance of aerial view images in ATR, we propose SCP-Label for post-processing, as illustrated in Fig.\ref{fig_intro}. To avoid the model being biased towards the majority classes, the proposed method leverages the re-sampled data as model input. In addition, in order to improve the fine-grained learning ability of the model, we utilize a backbone with an attention mechanism. Rather than directly assigning labels to each image, we use a clustering algorithm to cluster scenes, and then assign scene labels through model integration instead. 

In the experimental phase, we determine our chosen feature extraction backbone, data augmentation strategy, and under-sampling data volume through multiple experimental settings, and set up multiple ablation experiments to determine our key parameters. Both on Track 1 (SAR) and Track 2 (SAR+EO) of the CVPR 2022 PBVS Workshop MAVOC Challenge~\cite{mavoc}, our Top-1 accuracy rate in the final test phase reach \textbf{36.44\%} and \textbf{51.09\%}, respectively. Experimental results show that our method \textbf{performs the best among all participants}.

Our contributions can be summarized as follows:
\begin{itemize}
	\item We address the challenge of MAVOC by proposing a new pseudo-labeling strategy that can aggregate images of the same scene and assign pseudo-labels based on scene clustering.
	\item We propose a multi-modal integration scheme to generate pseudo-labels based on scene clustering. Compared with general pseudo-labeling, this design allows the model to have better generalization ability under different illumination, shooting angles and other disturbances.
	\item We achieve superior performance over the state-of-the-art approaches, and win first place on two tracks of the MAVOC challenge.
\end{itemize}
 

\section{Related work}
\label{sec2}

EO is the most popular sensor type in modern RS systems. It can efficiently capture images in the visible spectrum that are human-interpretable, and data collection for EO is fairly easy (for example, aerial images can be collected in the EO domain using the Google Earth~\cite{GoogleEarth} platform). In the past, ATR has been a lot of research work in the field of EO~\cite{lam2018xview,xia2018dota,long2021creating}. For the classification of EO images, traditional visual methods~\cite{arguedas2015texture} were used to extract texture features to classify ships at sea. Katherine~\cite{rice2018convolutional} used a VGG-16-based CNNs for detection and classification of EO images of ships at sea.

On the other hand, SAR data is also one of the sensor patterns for other commonly used ATR algorithms. Although less well understood, SAR has the advantage of operating at night and in different weather conditions, which gives it a distinct advantage over EO sensors in some applications. However, the acquisition process of SAR data is much more complex and costly, which means that the breadth of SAR-ATR studies lags behind EO-ATR studies in several ways~\cite{inkawhich2020advanced,inkawhich2021bridging,inkawhich2021training,chen2016target}. 

With the increasing application of deep learning in SAR, terrain surface classification~\cite{parikh2020classification}, resolution inversion~\cite{wang2016sea}, speckle removal~\cite{wang2017sar}, specific methods in interferometric SAR (InSAR) applications~\cite{anantrasirichai2018application}, SAR-optical data fusion~\cite{hughes2018identifying}, etc. have attracted extensive attention. Tzeng et al.~\cite{tzeng1998fuzzy} proposed a method for classifying SAR images using fuzzy logic and dynamic neural networks, which was an early attempt to classify SAR images using neural networks. Popular deep learning algorithms used end-to-end CNNs as feature extractors to generate discriminative features that can work with subsequent classifiers automatically~\cite{chen2014sar}. Geng et al.~\cite{geng2017deep} proposed a single polarimetric/supervised SAR image classification system to improve the problems of noise and speckle in the data that are not easy to characterize effectively.

There have been many outstanding studies of class imbalanced data in recent years. Important adjustment methods are mainly divided into re-sampling~\cite{buda2018systematic,byrd2019effect} and re-weighting~\cite{cui2019class,cao2019learning}. Re-sampling includes majority class~\cite{galar2013eusboost,liu2008exploratory} under-sampling or minority class~\cite{chawla2002smote,han2005borderline} over-sampling, which aims to rebalance the distribution of data. However, this approach may lead to overfit to the minority class while losing some valuable training samples, thereby deteriorating the modeling quality on large-scale datasets. Re-sampling weights, assigning higher weights to tails~\cite{cui2019class} or hard samples~\cite{cao2019learning} in the loss function, makes the expectation of each class closer to the test distribution. However, when applying re-weighting to large-scale real-world scenarios, it is often difficult to optimize. In addition, there are some other ways to mitigate the impact of model representation on long-tailed data, such as FocalLoss~\cite{lin2017focal}.

\section{How CNNs perform on MAVOC}
\label{sec3}

In this section, we attempt to investigate the misclassified and biased behavior of CNNs on MAVOC dataset. Many existing state-of-the-art image classification algorithms are trained on balanced and coarse-grained high-quality benchmarks, such as ImageNet~\cite{deng2009imagenet}, Cifar~\cite{krizhevsky2009learning}, etc. Furthermore, these benchmarks are all visible light images and the recognized objects generally have clean backgrounds, which is profitable for the model to capture features such as color and texture. In contrast, aerial view images present a unique set of challenges due to the fine-grained image features and long-tailed data distribution. In addition, other specific issues in aerial view images also pose further challenges such as the nature of the target, the ratio of the target size to the background of the image, lower resolution of the target, target texture, light reflectance, etc. Since the image properties of the two data domains are significantly different, we propose a conjecture on MAVOC dataset, the image classification model will be confused and biased due to high interclass similarity, skewed class distribution, which may further result in severe performance on validation and test set.

Instead of extending the protocol, which utilizes various class-imbalanced ratios to produce long-tailed versions of benchmark datasets, such as Cifar, MAVOC dataset has been split into train, validation and test set. In order to justify our conjecture, we observe and analyze the performance of official baseline, implemented with Resnet50~\cite{he2016deep}. Concretely, we manually divide the training set and analyze the classification of the baseline on the artificial validation set.

\begin{figure}[t]
	\centering
	\includegraphics[width=0.48\textwidth]{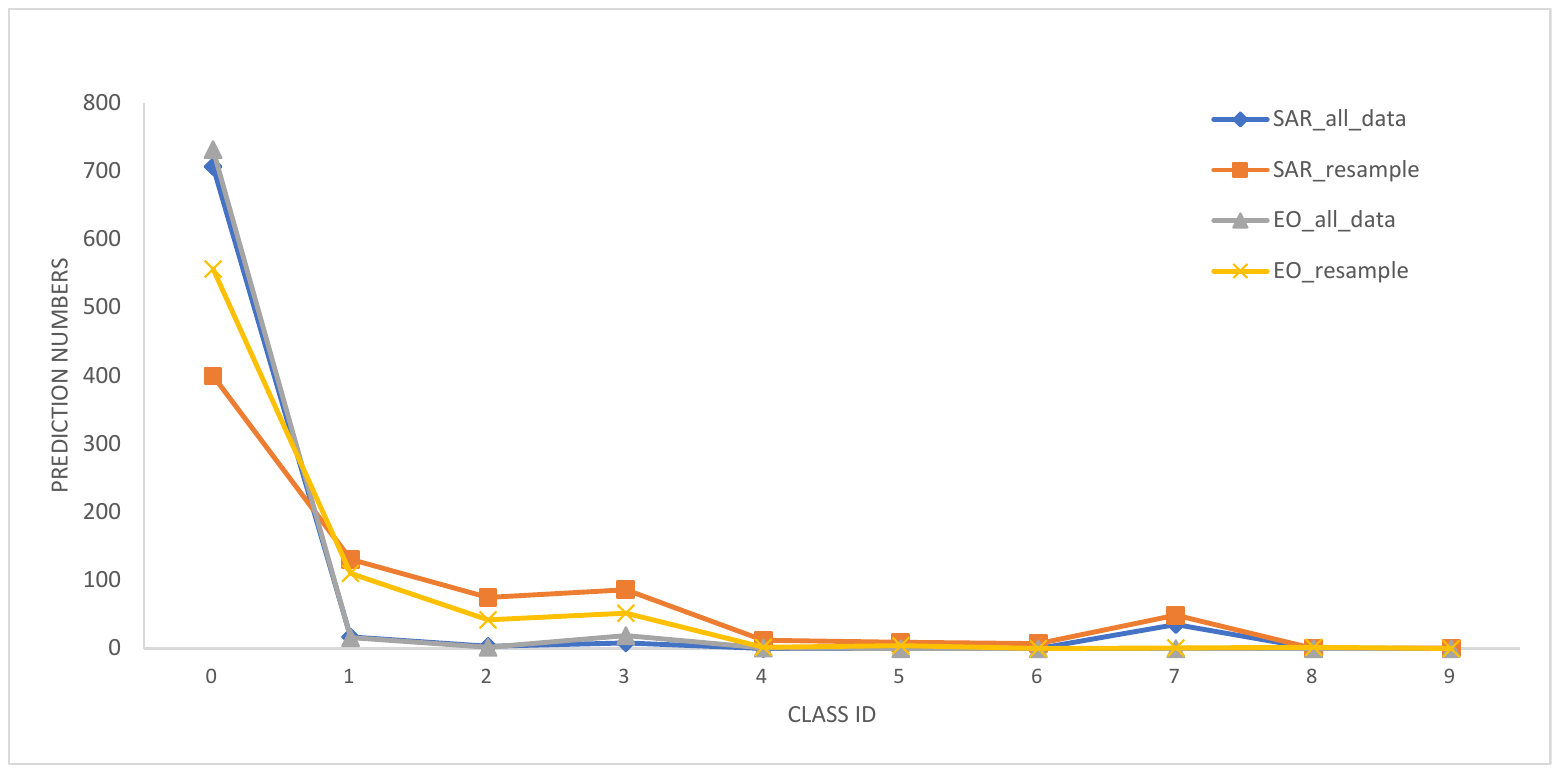} 
	\caption{The baseline is biased towards the majority class. Resampling will alleviate the class imbalance problem.}
	\label{fig_dataset_bias}
\end{figure}

Firstly, we manually divide the training set. Specifically, we sample 60 images from each category as the artificial validation set to keep the division result similar to the official validation set, and utilize the official baseline for training. As illustrated in Fig.\ref{fig_dataset_bias}. we can find that the long-tailed distribution of the training set causes the model to be biased towards the majority class both on SAR and EO set. Therefore, in order to solve the serious data imbalance, we re-sample 3000 images in each category for training. The accuracy rates before and after resampling show that resampling can effectively alleviate the data imbalance.

\begin{figure}[t]
	\centering
	\includegraphics[width=0.45\textwidth]{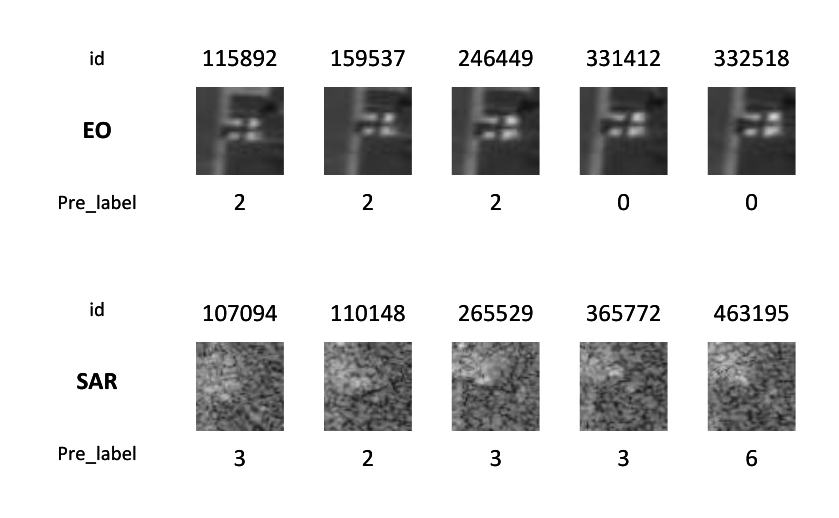} 
	\caption{Scene aggregation properties in EO and SAR images.}
	\label{fig_dataset_scence_cluster}
\end{figure}

Crucially, we find that the MAVOC dataset has scene aggregation characteristics, that is, the same target in the same scene exists multiple images with different sensors, shooting angles, blur degrees, and illumination. Due to the influence of the above multiple distractions, objects in the same scene are identified as different categories. As shown in Fig.\ref{fig_dataset_scence_cluster}, in a scene we selected, obviously, the results of all the images should be the same, whereas the performance of the classifier is not the case. Through observation, we can find that under the action of different factors, the model's focus is not the same. Therefore, if we can identify targets in the same scene as the same category through corresponding data augments or model integration methods, the classification accuracy will improve over baseline by a large margin.

Besides, the MAVOC challenge is not a multi-label classification task, although in some of the images, multiple targets exist at the same time. This finding also poses certain challenges for us, and we have enhanced the model's ability to learn to the corresponding labeled targets by means of data augments such as Cutmix, which mitigate the effect of multiple targets to some extent.

In order to solve the above problems, we introduce pseudo-labeling strategy based on scene clustering (SCP-Label) for post-processing.  

\section{Approach}
\label{sec4}

\subsection{Overall framework}

\begin{figure*}
	\centering
	\includegraphics[width=1.0\textwidth]{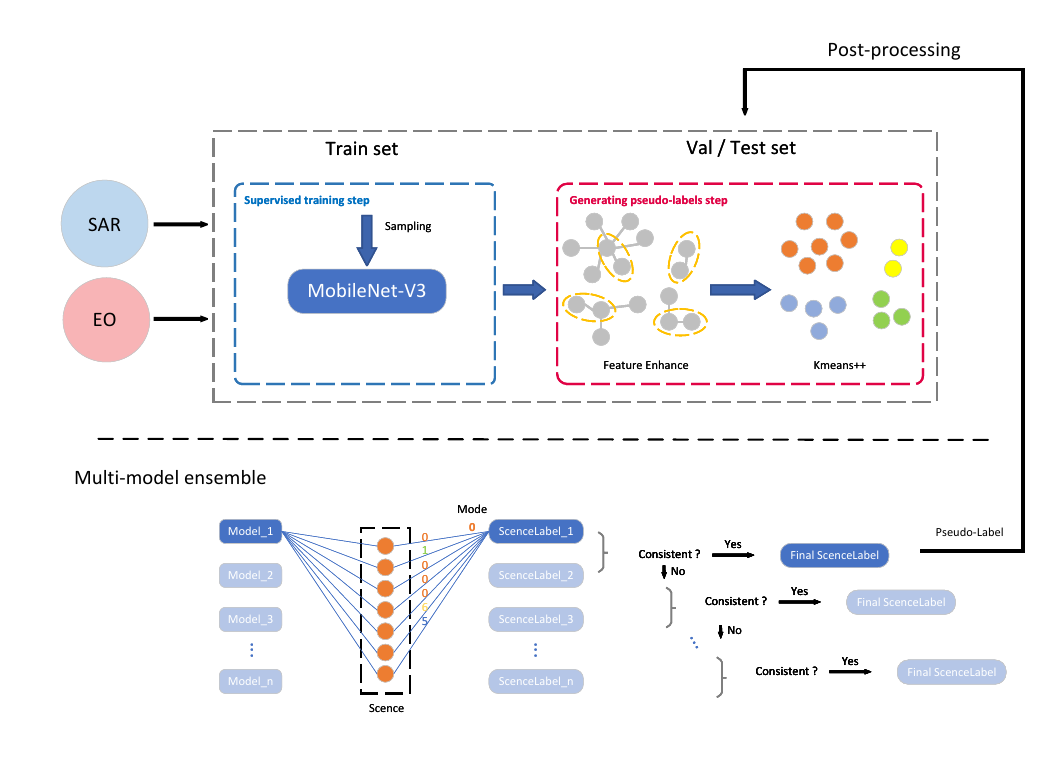} 
	\caption{{\bf Overall framework for SCP-Label.} The overall framework is divided into two steps, namely the supervised training step and the generating pseudo-labels stage. By extracting features with the pre-trained model, we utilize the DBA feature-enhanced features for scene clustering. And our model ensemble method leverages a strategy of comparing votes one by one.}
	\label{fig_framework}
\end{figure*}

As shown in Fig.\ref{fig_framework}, our SCP-Label consists of two training steps, supervised training step and generating pseudo-labels step.

\subsubsection{Supervised training step}\label{subsubsec411}

In this step, the model contains three main components. Concretely, we use MobileNet as our backbone, introduce SeNet as our attention module, and utilize global average pooling (GAP) for feature aggregation.

MobileNet~\cite{howard2017mobilenets} as a general backbone usually ignores the subtle but discriminative features in fine-grained recognition, hence a network structure with an attention mechanism is necessary. In order to satisfy the needs of fine-grained recognition, SeNet~\cite{hu2018squeeze} can effectively capture the fine-grained features rather than original structure.

Considering the labeled set, let $x$ denote a training sample with its corresponding label $y \in {1, 2,\cdots,C}$ for a C-class recognition task. For class-imbalanced distribution, we separately apply under-sampling and over-sampling for the majority and minority classes, respectively.

After sampling, two samples $(x_{s}, y_{s})$ and $(x_{e}, y_{e})$ are obtained as the input data, where $(x_{s}, y_{s})$ is from the SAR domain and $(x_{e}, y_{e})$ is from the EO domain. Then, the sampled data from the two domains are loaded into backbone respectively, and by GAP the feature vectors  $ \mathbf{f}_{s} \in \mathbb{R}^D$ and $ \mathbf{f}_{e} \in \mathbb{R}^D$ can be acquired.

During the training phase of the classifier, the predicted output  $z\in\mathbb{R}^C$ is illustrated as

\begin{equation}
	z=\mathbf{W}_{s}^ \top \mathbf{f}_{s}\quad or \quad\mathbf{W}_{e}^ \top \mathbf{f}_{e}
\end{equation}

Then softmax function will utilize each component in z, i.e., $[z_{1}, z_{2},\cdots,z_{C} ]^\top $, to calculate the probability for each category $i \in \{1, 2,\cdots,C\}$ by

\begin{equation}
	\hat{\mathbf{p}_{i}}=\frac{e^{z_{i}}}{\sum_{j=1}^{C}e^{z_{i}}}
\end{equation}%

Generally, we denote the output probability distribution as $\hat{\mathbf{p}} =[\hat{p_{1}}, \hat{p_{2}},\cdots, \hat{p_{C}} ]^\top$, $E(\cdot,\cdot)$ as the cross-entropy loss function. Therefore, our supervised training step generates a weighted cross-entropy recognition loss, which is formulated as

\begin{equation}
	L={E(\mathbf{\hat{\mathbf{p}}},y_{s}) }\quad or \quad{ E(\mathbf{\hat{\mathbf{p}}},y_{e})}
\end{equation}%

\subsubsection{Generating pseudo-labels step}\label{subsubsec412}

As observed in Sec.\ref{sec3}, the same target in the same scene will be identified as different categories due to various factors. Therefore, instead of per image strategy of pseudo label, we propose pseudo-labeling strategy based on scene clustering.

The converged model in the supervised training step will serve as a pre-trained model for generating pseudo-labels step. The feature vector after feature extraction by backbone and GAP will undergo L2-normalize and feature enhancement. As a commonly used feature enhancement strategy in retrieval, DataBase-side Augmentation (DBA)~\cite{arandjelovic2012three} particularly helps improve feature quality, where every feature in the database is replaced with a weighted sum of the point’s own value and those of its top k nearest neighbors (k-NN). Moreover, feature enhancement can further strengthen the scene aggregation characteristics and alleviate the intra-cluster distance. 

As an improved algorithm of Kmeans~\cite{hartigan1979algorithm} clustering, Kmeans++~\cite{arthur2006k} has the advantages in speed and accuracy, which is used as the clustering algorithm for our scene clustering. By specifying the initial number of clusters, we cluster the enhanced feature vectors and calculate the sum of squared distances within each cluster, discarding the clusters with poor quality.

Then, we assign cluster labels using a multi-model ensemble approach. The single model exploits the mode of the predicted labels of all images in the cluster as the cluster label. However, the model is prone to be affected by model bias and inter-class confusion, so we utilize one-by-one comparison for multi-model ensemble. Specifically, if the cluster labels of model\_1 and model\_2 are consistent, the label is used as the final cluster label, and the model ensemble is terminated. Otherwise it continues to compare model\_2 and model\_3. If the model ensemble does not terminate early, the comparison is made until the last model, and the cluster label of the last model is used as the final cluster label.

SCP-Label, as pseudo-label, will be applied directly to the validation set or test set as a post-processing method, so the black arrow "pseudo label" connects the top and the bottom parts. Since our pseudo label is only done as a post-processing and not as a semi-supervised learning, the label of each cluster will be directly output as a category without any subsequent task.

\subsection{Usage of SCP-Label}

SCP-Label can be used in two situations. On the one hand, it can be employed as a post-processing method for the test set, through which erroneous class labels within clusters can be corrected. On the other hand, it can be used as a pseudo-label strategy under the semi-supervised learning framework to achieve end-to-end learning, such as FixMatch~\cite{sohn2020fixmatch}.

\section{Experiments}
\label{sec5}

In this section, we first introduce the dataset and implementation details, and then quantitatively evaluate the performance of our method on the dataset. Finally, some ablation studies are performed to demonstrate the effectiveness of each component of the network. On both Track 1 (SAR) and Track 2 (SAR+EO) of CVPR 2022 PBVS Workshop MAVOC Challenge, our Top-1 Accuracy ranked 1st place in the final leaderboard.

\subsection{Dataset}

\begin{figure*}[t]
	\centering
	\begin{subfigure}{0.01\linewidth}
		\rotatebox{90}{\scriptsize{Electro-Optical}}
	\end{subfigure} 
	\begin{subfigure}{0.01\linewidth}
		\rotatebox{90}{\scriptsize{(EO)}}
	\end{subfigure} 
	\begin{subfigure}{0.092\linewidth}
		\centering
		\textbf{0}\\
		\includegraphics[width=.8\textwidth]{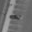}\\
		\includegraphics[width=.8\textwidth]{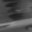}
	\end{subfigure}
	\begin{subfigure}{0.092\linewidth}
		\centering
		\textbf{1}\\
		\includegraphics[width=.8\textwidth]{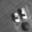}\\
		\includegraphics[width=.8\textwidth]{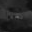}
	\end{subfigure}
	\begin{subfigure}{0.092\linewidth}
		\centering
		\textbf{2}\\
		\includegraphics[width=.8\textwidth]{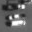}\\
		\includegraphics[width=.8\textwidth]{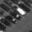}
	\end{subfigure}
	\begin{subfigure}{0.092\linewidth}
		\centering
		\textbf{3}\\
		\includegraphics[width=.8\textwidth]{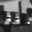}\\
		\includegraphics[width=.8\textwidth]{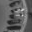}
	\end{subfigure}
	\begin{subfigure}{0.092\linewidth}
		\centering
		\textbf{4}\\
		\includegraphics[width=.8\textwidth]{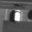}\\
		\includegraphics[width=.8\textwidth]{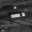}
	\end{subfigure}
	\begin{subfigure}{0.092\linewidth}
		\centering
		\textbf{5}\\
		\includegraphics[width=.8\textwidth]{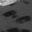}\\
		\includegraphics[width=.8\textwidth]{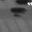}
	\end{subfigure}
	\begin{subfigure}{0.092\linewidth}
		\centering
		\textbf{6}\\
		\includegraphics[width=.8\textwidth]{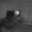}\\
		\includegraphics[width=.8\textwidth]{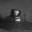}
	\end{subfigure}
	\begin{subfigure}{0.092\linewidth}
		\centering
		\textbf{7}\\
		\includegraphics[width=.8\textwidth]{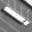}\\
		\includegraphics[width=.8\textwidth]{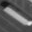}
	\end{subfigure} 
	\begin{subfigure}{0.092\linewidth}
		\centering
		\textbf{8}\\
		\includegraphics[width=.8\textwidth]{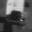}\\
		\includegraphics[width=.8\textwidth]{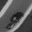}
	\end{subfigure} 
	\begin{subfigure}{0.092\linewidth}
		\centering
		\textbf{9}\\
		\includegraphics[width=.8\textwidth]{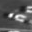}\\
		\includegraphics[width=.8\textwidth]{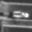}
	\end{subfigure} \\
	
	\vspace{0.05cm}
	\rule{.99\textwidth}{0.96pt}\\
	\vspace{0.05cm}
	
	\centering
	\begin{subfigure}{0.01\linewidth}
		\rotatebox{90}{\scriptsize{Synthetic Aperture Radar}}
	\end{subfigure} 
	\begin{subfigure}{0.01\linewidth}
		\rotatebox{90}{\scriptsize{(SAR)}}
	\end{subfigure} 
	\begin{subfigure}{0.092\linewidth}
		\centering
		\includegraphics[width=.8\textwidth]{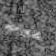}\\
		\includegraphics[width=.8\textwidth]{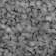}
	\end{subfigure}
	\begin{subfigure}{0.092\linewidth}
		\centering
		\includegraphics[width=.8\textwidth]{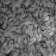}\\
		\includegraphics[width=.8\textwidth]{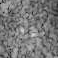}
	\end{subfigure}
	\begin{subfigure}{0.092\linewidth}
		\centering
		\includegraphics[width=.8\textwidth]{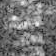}\\
		\includegraphics[width=.8\textwidth]{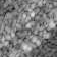}
	\end{subfigure}
	\begin{subfigure}{0.092\linewidth}
		\centering
		\includegraphics[width=.8\textwidth]{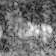}\\
		\includegraphics[width=.8\textwidth]{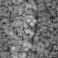}
	\end{subfigure}
	\begin{subfigure}{0.092\linewidth}
		\centering
		\includegraphics[width=.8\textwidth]{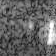}\\
		\includegraphics[width=.8\textwidth]{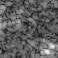}
	\end{subfigure}
	\begin{subfigure}{0.092\linewidth}
		\centering
		\includegraphics[width=.8\textwidth]{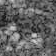}\\
		\includegraphics[width=.8\textwidth]{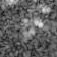}
	\end{subfigure}
	\begin{subfigure}{0.092\linewidth}
		\centering
		\includegraphics[width=.8\textwidth]{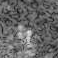}\\
		\includegraphics[width=.8\textwidth]{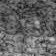}
	\end{subfigure}
	\begin{subfigure}{0.092\linewidth}
		\centering
		\includegraphics[width=.8\textwidth]{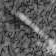}\\
		\includegraphics[width=.8\textwidth]{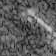}
	\end{subfigure}
	\begin{subfigure}{0.092\linewidth}
		\centering
		\includegraphics[width=.8\textwidth]{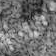}\\
		\includegraphics[width=.8\textwidth]{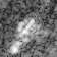}
	\end{subfigure}
	\begin{subfigure}{0.092\linewidth}
		\centering
		\includegraphics[width=.8\textwidth]{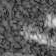}\\
		\includegraphics[width=.8\textwidth]{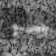}
	\end{subfigure} \\

	\caption{Two sample pairs of EO and SAR chips from each of the 10 classes in the Unicorn Dataset.}
	\label{fig:exp_fig1}
\end{figure*}

The data for this challenge consists of two types of small window regions (chips) generated from large images captured by several aircraft-mounted EO and SAR sensors. An EO chip is a 31 ${\times}$ 31 px image. A SAR chip covers the same approximate field of view as the corresponding EO image and has a finer resolution than the EO image. Chips vary in pixel size due to SAR processing, but are typically around 55 ${\times}$ 55 px. \cref{fig:exp_fig1} provides the samples of EO and SAR chips. The target belongs to a list of 10 classes that correspond to a training set with a non-uniform distribution of the number of samples per class, while the validation and test sets are based on a small number of uniformly distributed samples per class.

The dataset is divided into:
\begin{itemize}
	\item Training set: The data in this set is seriously imbalanced. (i.e. some classes have more samples than others)
	\item Validation set: This set is a uniformly distributed among all classes with $\le$ 100 samples per class.
	\item Test set: This split resembles the validation test with a uniform distribution of testing images among the classes.
\end{itemize}

These images belong to one of ten categories: 0 to 9. The train data contains SAR and EO images and class labels. Validation / test data consists of SAR images from Track 1 (SAR) as well as SAR and EO images from Track 2 (SAR+EO) of the challenge. The goal is to use the provided (SAR+EO) training images to maximize classification accuracy when the input is only SAR images or SAR and EO images. \cref{tab:exp_table1} shows a breakdown of images for each category. Note that now it's not a class name, but simply labeled 0-9.

\begin{table}
	\centering
	\caption{Details of the Unicorn Dataset used in this challenge (counts represent the number of (EO, SAR) pairs).}
	\begin{tabular}{@{}clccc@{}}
		\toprule
		Class  &  Class Name &  Train &  Val &  Test  \\
		\midrule
		0 &  sedan & 234,209& -&-\\
		1 &  suv & 28,089& -&-\\
		2 &  pickup truck&  15,301& -&-\\
		3 & van&  10,655& -&-\\
		4 & box truck&  1,741& -&-\\
		5 & motorcycle& 852& -&-\\
		6 & flatbed truck&  828& -&-\\
		7 & bus&  624& -&-\\
		8 & pickup truck with trailer&  840& -&-\\
		9 & flatbed truck with trailer& 633& -&-\\
		\midrule
		& Total& 293,772& 770& 826\\
		\bottomrule
	\end{tabular}
	\label{tab:exp_table1}
\end{table}

\subsection{Implementation details}

For the methods we proposed, we under-sample the classes with data volume greater than 1741 to 1741. We use the stratified sampling to divide the dataset on development phase, randomly selecting 70\% of each category as the training set and 30\% as the validation set, so as to adjust training strategies and parameters offline. But on testing phase, we no longer partition data sets. In the stage of data preprocessing, we resize all EO image to 104 ${\times}$ 104. While the resolution of SAR image remain unchanged, we use 0 to fill its edge to 64 ${\times}$ 64. During the training, we randomly crop all SAR image to 56 $\times$ 56, augment the data by Rotation and Flipping and use Cutmix~\cite{yun2019cutmix} to reduce noise from the input image.

As shown in \cref{tab:exp_table2}, for the serious class imbalance problem, we design a total of 7 models, including three for Track 1 (1, 2, 3) and four for Track 2 (4, 5, 6, 7). Their differences are mainly reflected in training data and training epoch. We adopt MobileNetV3-large~\cite{howard2019searching} as feature extractor. In addition, we use an SGD optimizer with 0.9 momentum and ${1e^{-3}}$ weight decay, as well as fix learning rate ${1e^{-3}}$. Our model is implemented using PyTorch with 1 NVIDIA A100 GPU. Experiments are also conducted using MindSpore.

\begin{table}
	\centering
	\caption{Details of the 7 models we designed for Track 1 and 2. Model 1, 2 and 3 are for Track 1, and model 4, 5, 6 and 7 are for Track 2.}
	\scalebox{0.85}{
	\begin{tabular}{@{}ccrcc@{}}
		\toprule
		Models  &  Type &  Input classes &  Epochs  & Top-1 Acc (\%)\\
		\midrule
		1 &  SAR & [0,1,2,3,4,5,6,7,8,9]& 3  & 29.78\\
		2 &  SAR&  [4,5,6,7,8,9]& 20 & - \\
		3 &  SAR&  [6,7,8,9]& 30 & -\\
		4 &  EO&   [0,1,2,3,4,5,6,7,8,9]& 80 & 35.47\\
		5 &  EO&   [1,2,3,4,5,6,7,8,9]& 70& -\\
		6 &  EO&   [4,5,6,7,8,9]& 70& -\\
		7 &  SAR&  [4,6,8,9]& 20& -\\
		\bottomrule
	\end{tabular}
	\label{tab:exp_table2}
	}
\end{table}

\begin{table}[t]
	\centering
	\caption{The effect of EO images of different sizes on the accuracy on validation set.}
	\scalebox{0.9}{
		\begin{tabular}{@{}c|c|c|c@{}}
			\toprule
			Size &    Top-1 Accuracy &  Size &    Top-1 Accuracy (\%)  \\ 
			\midrule
			
			${32 \times 32}$& 15.19 & ${104 \times 104}$& \textbf{23.77} \\
			${56 \times 56}$& 19.61 & ${112 \times 112}$& 17.53 \\
			${64\times 64}$& 22.34 & ${120 \times 120}$& 18.05 \\
			${96 \times 96}$& 22.21 & ${128 \times 128}$& 17.27 \\
			
			\bottomrule
		\end{tabular}
	}
	\label{tab:exp_table4}
\end{table}

\begin{table*}[t]
	\centering
	\caption{Validation results for the CVPR 2022 PBVS Workshop MAVOC Challenge Track 1 with different setting.}
	\label{table1}
	\scalebox{0.95}{
	\begin{tabular}{|c|c|c|c|c|c|c|c|c|}
		\hline
		\multicolumn{1}{|c|}{\multirow{2}{*}{Method}} &\multirow{2}{*}{Backbone}&\multirow{2}{*}{Under-sample} & \multicolumn{5}{c|}{Augmentation Method} &\multirow{2}{*}{Top-1 Accuracy (\%)}\\
		\cline{4-8}
		
		\multicolumn{1}{|c|}{} & & & Rotation & Flipping & RA & TA & Cutmix &  \\
		\hline
		
		\multicolumn{1}{|c|}{\multirow{11}{*}{Single model}} &\multirow{4}{*}{Resnet50}&\multirow{4}{*}{all data} & \checkmark & \checkmark & - & - & - &\multicolumn{1}{c|}{ 16.10}\\
		\cline{4-9}
		
		\multicolumn{1}{|c|}{} & & & - & - & \checkmark & - & - & 15.84 \\
		\cline{4-9}
		\multicolumn{1}{|c|}{} & & & - & - & - & \checkmark & - & 16.10  \\
		\cline{4-9}
		\multicolumn{1}{|c|}{} & & & \checkmark & \checkmark & - & - & \checkmark & \textbf{17.10}  \\
		\cline{2-9}
		\multicolumn{1}{|c|}{} & Efficientnet-b1& all data& \checkmark & \checkmark & - & - & \checkmark & 15.88  \\
		\cline{2-9}
		\multicolumn{1}{|c|}{} & Swin-Transformer& all data& \checkmark & \checkmark & - & - & \checkmark & 16.88  \\
		\cline{2-9}
		\multicolumn{1}{|c|}{} & DenseNet161& all data& \checkmark & \checkmark & - & - & \checkmark & 18.05  \\
		\cline{2-9}
		\multicolumn{1}{|c|}{} & {\multirow{4}{*}{MobileNetV3-large}}& all data& \checkmark & \checkmark & - & - & \checkmark & \textbf{18.18}  \\
		\cline{3-9}
		\multicolumn{1}{|c|}{} & & 4000& \checkmark & \checkmark & - & - & \checkmark & 19.35 \\
		\cline{3-9}
		\multicolumn{1}{|c|}{} & & 2000& \checkmark & \checkmark & - & - & \checkmark & 19.74 \\
		\cline{3-9}
		\multicolumn{1}{|c|}{} & & 1741& \checkmark & \checkmark & - & - & \checkmark & \textbf{21.30} \\
		\hline
		
	\end{tabular}
	}
	\label{tab:exp_table3}
\end{table*}

\begin{table}[h]
	\centering
	\caption{ Latest update of the test results for the PBVS 2022 Multi-modal Aerial View Object Classification Challenge Track 1 (SAR) and Track 2 (SAR+EO).}
	\scalebox{0.9}{
	\begin{tabular}{@{}c|c|c@{}}
		\toprule
		Rank &    Top-1 Acc of Track 1 (\%) &  Top-1 Acc of Track 2 (\%)  \\ 
		\midrule
		1st &  \textbf{36.44} (Ours)&  \textbf{51.09} (Ours)\\
		2nd & 31.23& 46.85\\
		3rd & 28.09& 41.77\\
		4th & 27.97& 37.65\\
		5th & 27.48& 34.26\\
		\bottomrule
	\end{tabular}
	}
	\label{tab:exp_table6}
\end{table}

In terms of feature enhancement, the k1 coefficient of DBA is 1. In the Kmeans++ algorithm for scene clustering, we choose the K value to be 80.

\subsection{Main results}

\cref{tab:exp_table3} shows the experimental results on validation set of Track 1 (SAR), including the results from different backbone and data augmentation strategies. It can be seen that when MobileNetV3-large is used to extract features and the number of under-samples is 1741, the data augmentation effect is the best when when using Rotation, Flipping and Cutmix. In addition, we also introduce mainstream data augmentation methods, such as RandAugment(RA)~\cite{cubuk2020randaugment} and TrivialAugment(TA)~\cite{muller2021trivialaugment}, but they do not perform well in this dataset, and we think that the use of too many augmentation operations cause the image to lose a lot of original information.

Since EO and SAR images are matched and the category distribution is consistent, we also apply the same backbone, under-sampling data amount and data augmentation to the EO dataset. In addition, since the size of EO images will affect the feature effect of CNNs extraction, we conduct tests on EO with different sizes, as shown in \cref{tab:exp_table4}. When the input size of the EO image is 104 ${\times}$ 104, the test effect on the validation set is the best, which is nearly 9\% higher than the original input size of 32 ${\times}$ 32.

The highest score of our single model (the result of testing on the test set) is shown in \cref{tab:exp_table2}, where a single model Top-1 Accuracy using only SAR image data can reach 29.78\%, while using only EO image data The Top-1 Accuracy of the first model can reach 35.47\%, and the two Top-1 Accuracies can achieve the top 5 results in Track 1 (SAR) and Track 2 (SAR+EO) respectively. Compared with the best results of a single model, after using our multi-model ensemble SCP-Label method, we improve the Top-1 Accuracy by 6.66\% and 15.62\% on the two tracks, respectively, and \textbf{achieve the state-of-the-art results on both tracks}, as shown in \cref{tab:exp_table6}.

The top 5 teams from the latest update of the test results for the CVPR 2022 PBVS Workshop MAVOC Challenge Track 1 (SAR) and Track 2 (SAR+EO) are listed in terms of the top-1 accuracy in the testing phase in \cref{tab:exp_table6}. It can be seen that we won the first place on both tracks. On both tracks, our accuracy is 36.44\% and 51.09\% respectively, far ahead of the second place.

\begin{table}[t]
	\centering
	\caption{The effect of different K values on the Kmeans clustering effect.}
	\begin{tabular}{@{}c|c|c|c@{}}
		\toprule
		K &    Inertias &  Silhouette &Calinski-Harabaz Index \\ 
		\midrule
		20 & 227& 0.25 & 48\\
		40 & 167& 0.23 & 37\\
		60 & 136& 0.25 & 32\\
		80 & 113& 0.26 & 30\\
		100 & 98& 0.26 & 27\\
		120 & 84& 0.26 & 27\\
		140 & 74& 0.27 & 26\\
		\bottomrule
	\end{tabular}
	\label{tab:exp_table7}
\end{table}

\begin{table}[t]
	\centering
	\caption{The influence of the k1 coefficient of different DBAs on the clustering effect.}
	\begin{tabular}{@{}c|c|c|c@{}}
		\toprule
		k1 &    Inertias &  Silhouette &Calinski-Harabaz Index \\ 
		\midrule
		1 & 39& 0.50 & 93\\
		2 & 53& 0.45 & 68\\
		3 & 73& 0.38 & 48\\
		\bottomrule
	\end{tabular}
	\label{tab:exp_table8}
\end{table}

\subsection{Ablation studies}

Kmeans++ is one of the most commonly used methods in clustering, which calculates the best class attribution based on the similarity of point-to-point distances. Among them, the choice of K value is particularly important. We use a variety of clustering-related evaluation indicators to choose the appropriate K value, including Inertias (it represents the sum of the distances from the sample to the nearest cluster center. The smaller its value, the better, indicating that the distribution of samples between classes is more concentrated) , Silhouette (the higher the value, the better, the closer the distance between samples of the same class and the farther the distance between samples of different classes), and the Calinski-Harabaz index (for K clusters, the Calinski-Harabazindex defined as the ratio of between-group dispersion to within-group dispersion, the larger the score, the better the clustering effect). As shown in \cref{tab:exp_table7}, we record the feedback for each metric after trying Kmeans clustering with different values of K. Through the feedback of the results of the three indicators and our attempts in this challenge, we determin that the K value of Kmeans++ is 80.

DBA is the feature enhancement on the database side, which only uses the attributes of the adjacent images on the database side to expand the features of the image itself, thereby enhancing the features. After we have determined the K value of the clustering, we can further improve the clustering effect by enhancing the features. As shown in \cref{tab:exp_table8}, we record the effect of the DBA k1 coefficient on various metrics of Kmeans++. Compared with k1 = 2 or k1 = 3, when k1 = 1, the smaller the Inertias, the larger the Silhouette and Calinski-Harabaz Index, indicating that the kmeans++ clustering effect is better. Therefore, we determine the k1 coefficient of DBA to be 1.


\section{Conclusion}
\label{sec6}

In this work, we propose a multi-model ensemble pseudo-label strategy based on scene clustering, named SCP-Label, to improve the Top-1 accuracy of predicting SAR and EO test set images. The motivation for this strategy is that we observe that this dataset not only has the problem of extreme imbalance of samples between classes, but also has the problem of fine grain within classes, and has scene aggregation characteristics. SCP-Label performs scene clustering on the test set, and then uses multiple models trained from different angles to label each sample in each cluster with pseudo-labels, and finally obtains the cluster label of each cluster, and cluster labels are mapped to each sample in the cluster to obtain the final output. Our method \textbf{win champions on both Track 1 (SAR) and Track 2 (SAR+EO)} of the CVPR 2022 PBVS Workshop MAVOC Challenge.

\section{Acknowledgement}

This work is sponsored by Natural Science Foundation of China(62276242), CAAI-Huawei MindSpore Open Fund(CAAIXSJLJJ-2021-016B), Anhui Province Key Research and Development Program(202104a05020007), and USTC Research Funds of the Double First-Class Initiative(YD2350002001)

{\small
\bibliographystyle{ieee_fullname}
\bibliography{egbib}
}

\end{document}